\documentclass{article}
\usepackage{ijcai24}

\usepackage{times}
\usepackage{soul}
\usepackage{url}
\usepackage[hidelinks]{hyperref}
\usepackage[utf8]{inputenc}
\usepackage[small]{caption}
\usepackage{graphicx}
\usepackage{amsmath}
\usepackage{amsthm}
\usepackage{booktabs}
\usepackage{algorithm}
\usepackage{algorithmic}
\usepackage[switch]{lineno}

\usepackage{bm}
\usepackage{amssymb}
\usepackage[table]{xcolor}
\usepackage{pgfplots}
\usepackage{subcaption}
\usepackage{tikz}
\usepackage{multirow}

\pgfplotsset{compat=newest}

\definecolor{mygray}{gray}{.9}
\definecolor{sota_blue}{HTML}{0071bc}
\makeatletter
\newcommand{\thickhline}{%
	\noalign {\ifnum 0=`}\fi \hrule height 1pt
	\futurelet \reserved@a \@xhline
}
\makeatother
\newcommand{\tablestyle}[2]{\setlength{\tabcolsep}{#1}\renewcommand{\arraystretch}{#2}\centering\footnotesize}

\title{State-space Decomposition Model for Video Prediction Considering Long-term Motion Trend}

\author{
Fei Cui$^{1}$\and
Jiaojiao Fang$^{1}$\and
Xiaojiang Wu$^1$\and
Zelong Lai$^{1}$\and 
Mengke Yang$^1$\and \\
Menghan Jia$^{1}$\And
Guizhong Liu$^1$\thanks{Corresponding author}
\affiliations
$^1$Xi'an Jiaotong University\\
}

\begin{document}

\maketitle

\begin{abstract}
    Stochastic video prediction enables the consideration of uncertainty in future motion, thereby providing a better reflection of the dynamic nature of the environment. Stochastic video prediction methods based on image auto-regressive recurrent models need to feed their predictions back into the latent space. Conversely, the state-space models, which decouple frame synthesis and temporal prediction, proves to be more efficient. However, inferring long-term temporal information about motion and generalizing to dynamic scenarios under non-stationary assumptions remains an unresolved challenge. In this paper, we propose a state-space decomposition stochastic video prediction model that decomposes the overall video frame generation into deterministic appearance prediction and stochastic motion prediction. Through adaptive decomposition, the model's generalization capability to dynamic scenarios is enhanced. In the context of motion prediction, obtaining a prior on the long-term trend of future motion is crucial. Thus, in the stochastic motion prediction branch, we infer the long-term motion trend from conditional frames to guide the generation of future frames that exhibit high consistency with the conditional frames. Experimental results demonstrate that our model outperforms baselines on multiple datasets.
\end{abstract}

\section{Introduction}
\label{sec:intro}
Video prediction involves capturing the implicit environmental dynamics embedded in videos, aligning with the prior knowledge of model-based reinforcement learning. Therefore, reasonable predictions about the future from conditional frames have many applications in decision tasks ~\cite{finn2017deep,piergiovanni2019learning,dugas2022navdreams}. Video prediction aims to capture the dynamic representation of the world by modeling the prior knowledge of how the environment operates. Given the inherent stochastic nature of the world ~\cite{denton2018stochastic}, deterministic approaches for video prediction ~\cite{wang2017predrnn,jin2020exploring,wu2020future,gao2019disentangling} fall short of capturing the complete dynamics of the environment. On the contrary, stochastic video prediction ~\cite{denton2018stochastic,franceschi2020stochastic,akan2021slamp}, not relying on deterministic generation rules, exhibits superior generalization ability.

In stochastic video prediction, the key lies in how to capture the implicit motion cues embedded in the video. In contrast to the background (such as static room layout and indoor furniture) exhibiting shifts with camera movements, the complexity of motion subjects (such as pedestrians and moving cars) is higher and characterized by randomness. Traditional motion predictions often adopt deterministic approaches to forecast changes in motion, neglecting the various plausible possibilities of future motion. Alternatively, some predictions assume that the background is stationary, limiting the applicability of prediction models in fields such as navigation and autonomous driving. SLAMP ~\cite{akan2021slamp} decomposes stochastic video prediction into appearance-motion components but does not consider the long-term history of motion. Our insight is that the future development of motion has stochasticity, while backgrounds, such as static room layout, furniture, etc., exhibit deterministic shifts over time. Predicting the movement of dynamic subjects like pedestrians, who possess their independent consciousness, presents a challenging task. Therefore, we propose to decompose the overall video prediction into deterministic appearance prediction and stochastic motion prediction, aiming to adaptively focus on challenging-to-predict parts of the dynamics and sample future motion possibilities from the predicted distribution. Similar to SRVP ~\cite{franceschi2020stochastic}, our stochastic motion prediction branch relies on learning the residual updates of the latent states for stochastic variables to learn the system's temporal evolution. Deterministic appearance prediction, on the other hand, involves determining the background's temporal evolution based on deterministic residual updates.

\begin{figure}[!t]
  \begin{center}
     \includegraphics[width=\linewidth]{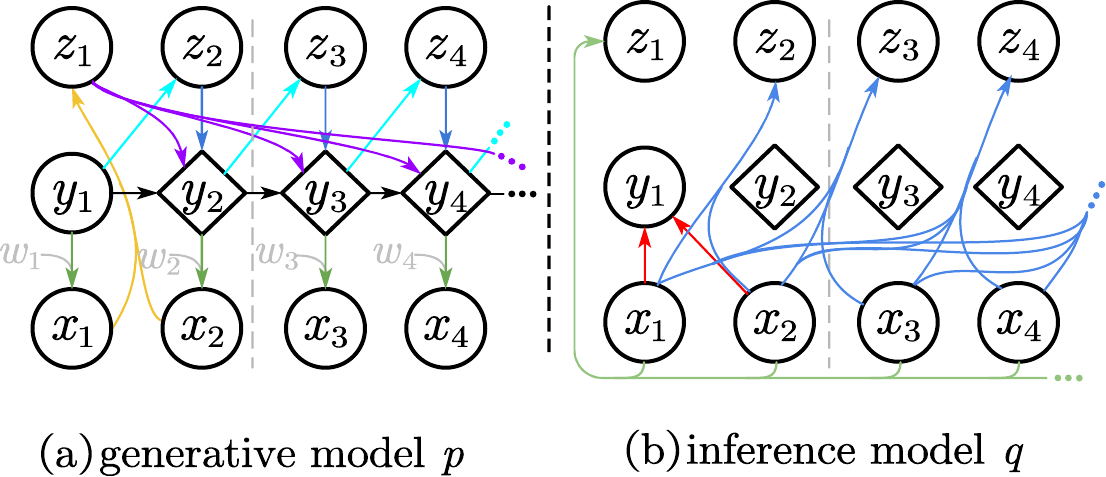}
  \end{center}
     \caption{Generative model $p$ and Inference model $q$ of our method, where circles and diamonds represent stochastic and deterministic variables, respectively. (a) In the generative model, the global motion trend variable $\bm{z}_1$ is generated from the conditional frames $\bm{x}_{1:k}$ (here $k=2$), and the local dynamic variable $\bm{z}_t$ is generated from the previous motion variable $\bm{y}_{t-1}$. (b) In the inference model, $\bm{z}_1$ is inferred from the complete sequence $\bm{x}_{1:T}$, and the local dynamic $\bm{z}_t$ is inferred from the frame sequence $\bm{x}_{1:t}$. The motion variable $\bm{y}_t$ and the appearance variable $\bm{w}_t$ are jointly decoded to generate the frame $\hat{\bm{x}_t}$.}
  \label{fig:generativeAndInference}
\end{figure}

In videos, there is implicit long-term historical information about motion. We aim to predict reasonable future frames based on the given conditional frames. Humans often make reasonable predictions about the future based on a few given frames because they match the temporally historical information inferred from the conditional frames with their long-term memory, allowing them to anticipate what will happen next. Therefore, inferring long-term motion trend from conditional frames is crucial for predicting the future. To achieve this, we infer the overall long-term motion trend from the complete input sequence as the global dynamic to assist in predicting the inner-frame transition in stochastic motion prediction branch.

Our contributions are summarized as follows:
\begin{itemize}
    \item We propose a state-space decomposition video prediction model that decomposes frame prediction into stochastic motion prediction and deterministic appearance prediction. Building upon a Gaussian prior for motion variables, the deterministic appearance prediction branch adaptively focuses on static features in frames.
    \item Our model utilizes a temporal transformer to infer the prior of global dynamics from the conditional frames, approximating the long-term motion trend. This results in the generation of future long-term sequence consistent with the ground truth.
    \item Experimental results demonstrate that the proposed approach achieves state-of-the-art performance on multiple datasets for the task of stochastic video prediction.
\end{itemize}  

\begin{figure*}[t]
  \begin{center}
     \includegraphics[width=\linewidth]{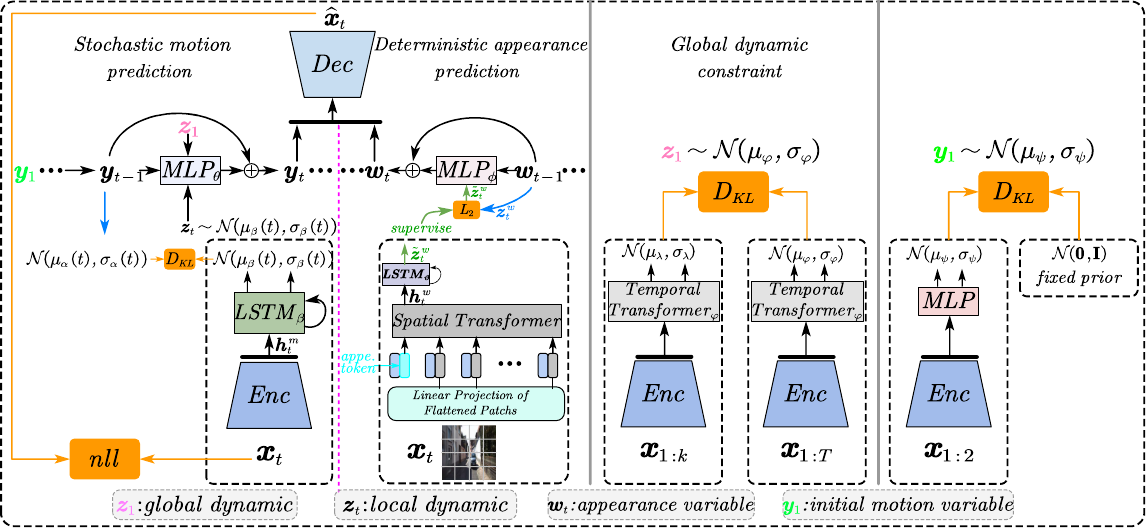}
  \end{center}
     \caption{Framework of our method. The original frames $\bm{x}_{1:T}$ are mapped to a latent space through an encoder, and a LSTM captures the temporal dynamics within this latent space in the motion prediction branch. In the appearance prediction branch, a ViT is employed to encode the static features related to the background. To encourage the motion varibales to disregard static features, a standard Gaussian prior is applied to the motion variables (\textbf{right}). The prior and posterior of the global dynamic variable $\bm{z}_1$ are inferred from the conditional frames $\bm{x}_{1:k}$ and input frames $\bm{x}_{1:T}$, respectively (\textbf{middle}).
     The frame $\hat{\bm{x}_t}$ is jointly decoded from the appearance variable $\bm{w}_t$ and the motion variable $\bm{y}_t$ (\textbf{left}). The training pipeline and testing pipeline are detailed in Appendix B.}
  \label{fig:framework}
\end{figure*}

\section{Related Work}

\subsection{Future Frame Video Prediction}
Video prediction aims to predict future frames from a few conditional frames, requiring the extraction of reasonable motion clues from the image frames. The temporal relationship implicit in the sequence of image frames is a focal point in the prediction task. Previous works have predominantly modeled the temporal dynamics from raw pixel frames ~\cite{vondrick2017generating,chatterjee2021hierarchical,ye2023video} or optical flow ~\cite{walker2016uncertain,liang2017dual,gao2019disentangling,akan2021slamp}. Recurrent neural networks ~\cite{wang2018eidetic,jin2020exploring} or transformers ~\cite{ye2023video,farazi2021local} have been commonly employed to infer motion in prior works. We also employ a recurrent model to infer frame-to-frame changes. The generation of video prediction can be deterministic ~\cite{xu2018structure,chang2022strpm}, but it struggles to capture the uncertainties present in the real world. Modeling the randomness of the future through latent variables and sampling from the predicted future distribution to generate upcoming image frames ~\cite{denton2018stochastic,franceschi2020stochastic} aligns more closely with real-world physical priors. Therefore, our work also captures complex future motion trends by predicting the distribution of future motion variables. This encompasses both local motion trends (frame-to-frame dynamics) and long-term motion trend (overall motion types, such as running, dancing, etc.).

\subsection{State-space Models}
The state-space model excels in modeling temporal sequences in a low-dimensional latent space ~\cite{hafner2019learning,karl2016deep,gregor2018temporal,franceschi2020stochastic,goel2022s,newman2023state}. Unlike many auto-regressive models ~\cite{weissenborn2019scaling,kalchbrenner2017video,micheli2022transformers,denton2018stochastic,akan2021slamp} commonly employed in prediction tasks, the temporal sequence in a state-space model flows in the latent space, decoupling the tasks of temporal prediction and frame generation. Consequently, the state-space model reduces the dependence on the encoder for generating the next frame. The state-of-the-art state-space model SRVP ~\cite{franceschi2020stochastic} models future sequence by predicting frame-to-frame residual terms, generating dynamically continuous sequences of future images. However, its content variables are inferred solely from the initial frame, limiting its applicability to tasks with a static background. In our approach, we also adopt a state-space model but predict the appearance variables that evolve over time, enabling adaptation to complex and dynamic scenarios. SLAMP ~\cite{akan2021slamp} decouples motion and appearance prediction, explicitly modeling the local motion history (optical flow) to predict the next frame. However, it does not consider the long-term motion trend. In our stochastic motion prediction, we introduce global motion constraints aimed at guiding the generation of future long-term sequence consistent with long-term motion trend.

\section{Method}


\subsection{Stochastic Motion Prediction}
The image frame sequence contains the dynamic features of moving subjects (such as motion type and speed of subjects) as well as the static features of the background (such as streets, trees, etc.). We define motion variable $\bm{y}$ and appearance variable $\bm{w}$, representing dynamic features related to the moving subject and static features associated with the background in the image frames, respectively. Consequently, at time step $t$, the original pixel frame $\bm{x}_t$ can be decoded using the motion variable $\bm{y}_t$ and the appearance variable $\bm{w}_t$ jointly. The image frame sequence $\bm{x}_{1:T}$ is encoded to derive the motion variables $\bm{y}_{1:T}$. Following the setup of the state-space model, the video dynamics evolve over time in the latent space of the motion variables. At time step $t$, the next motion variable $\bm{y}_{t+1}$ explicitly depends on current motion variable $\bm{y}_t$. Adhering to physical priors, the the motion of moving subject not only needs to adhere to the current local motion trend (i.e., inter frame local dynamic) but also must align with the long-term motion trend (i.e., global motion context). For instance, predicting long-term future frames under the condition of person running frames requires alignment with the person's long-term running context. Therefore, we define variable $\bm{z}_t$ ($2\le t\le T$) and $\bm{z}_1$ to represent the inter-frame local dynamic of frame $x_t$ with respect to $x_{t-1}$ and the global motion trend embedded in the video, respectively.
This temporal evolution of video dynamics is modeled using a residual network (ie., $\mathrm{MLP}_{\theta}$) to encapsulate the frame-to-frame transition as follows:
\begin{equation}
    \bm{y}_{t+1} =\bm{y}_t+\mathrm{MLP}_{\theta}(\bm{y}_t, \bm{z}_{t+1}, \bm{z}_1)
\label{equa: infer motion varibale}
\end{equation}
The overall generative model is illustrated in Figure \ref{fig:generativeAndInference}(a). At time step $t$, The local motion trend $\bm{z}_{t+1}$ is generated by the motion variable $\bm{y}_t$, denoted as $\bm{z}_{t+1}\sim \mathcal{N} (\mu _{\alpha} (\bm{y}_t), \sigma _{\alpha}(\bm{y}_t) )$. Simultaneously, the global  $\bm{z}_1$ is generated from the entire sequence of conditional frames $\bm{x}_{1:k}$, indicated by $\bm{z}_{1}\sim \mathcal{N} (\mu _{\lambda} (\bm{y}_{1:k}), \sigma _{\lambda} (\bm{y}_{1:k}) )$, where $k$ is the length of the conditional frames. To encourage motion variables to focus on dynamic features of the moving subject, we apply a standard Gaussian prior to the initial motion variable $y_1$ to discard unnecessary information, i.e, $\bm{y}_1\sim\mathcal{N}(0,I) $. The pixel frame $\bm{x}_t$ are generated by both the motion variable $\bm{y}_t$ and the appearance variable $\bm{w}_t$, expressed as $ \hat{\bm{x}}_t = Dec([\bm{w}_t, \bm{y}_t])$. 

\subsection{Deterministic Appearance Prediction}

In video prediction, moving subjects such as pedestrians and vehicles exhibit intricate motion patterns, reflecting the uncertainties present in the real world. Conversely, static objects in video streams, like room layouts, streets, and trees, remain stationary and may appear shift due to camera motion. Some existing works ~\cite{denton2017unsupervised,franceschi2020stochastic} separate the inference of content variables from motion prediction, but the assumption of invariant content variables constrains their performance in dynamic scenarios. Our insight is that static features associated with the background (such as the position, appearance, spatial relationships of static objects) undergo deterministic shifts over time, relying on the shift of the background. Hence, we define the variable $\bm{z}_{t+1}^w$ to represent inter-frame transition dynamic of appearance feature $\bm{w}_{t+1}$ with respect to $\bm{w}_{t}$. An MLP is used to predict appearance variables evolving over time.
\begin{equation}
    \bm{w}_{t+1}=\bm{w}_t+\mathrm{MLP}_{\phi}(\bm{w}_t, \bm{z}_{t+1}^{w} )
\label{equa: infer appearance variable}
\end{equation}
Unlike the prediction of stochastic distributions of motion variables, appearance variables transit deterministically. In other words, $\bm{z}_{t+1}^{w}$ is directly predicted from the previous appearance variable rather than sampled from a predicted distribution.  Additionally, in contrast to the strong Gaussian prior for initial motion variable $\bm{y}_1$, the initial appearance variable $\bm{w}_1$ is directly encoded from the initial frame $\bm{x}_1$.

The results from SRVP \cite{franceschi2020stochastic} have demonstrate that the strong Gaussian prior for initial motion variables encourages the motion variables to focus on the complex motion of the moving subjects, thereby neglecting unnecessary information for the motion of subjects. We employ a Vision Transformer (ViT) to encode appearance variables from pixel frames, as detailed in Section 3.4. The learnable appearance token encourages the ViT encoder to adaptively focus on some static background information in the pixel frames.

\subsection{Variational Inference}
After deriving the appearance variables $\bm{w}$, the complete evidence lower bound (ELBO) of the model can be derived. The conditional joint probability corresponding to the generative graph model shown in Figure~\ref{fig:generativeAndInference}(a) is given by:
\begin{equation}
    \begin{split}
        &p(\bm{x}_{1:T},\bm{z}_{1:T},\bm{y}_{1:T}|\bm{w}_{1:T})=p(\bm{z}_1|\bm{x}_{1:k})p(\bm{y}_1)\\
        & \textstyle\prod_{t=2}^{T}p(\bm{z}_t|\bm{y}_{t-1})p(\bm{y}_t|\bm{y}_{t-1},\bm{z}_t,\bm{z}_1) \textstyle\prod_{t=1}^{T}p(\bm{x}_t|\bm{y}_t,\bm{w}_t) 
    \end{split}
\label{equa: conditional joint probability}
\end{equation}
It can be observed that the conditional joint probability depends on the initial motion variable $\bm{y}_1$, the global dynamic variable $\bm{z}_1$, and the local dynamic variables $\bm{z}_{2:T}$. 

The latent variable $\bm{z}_1$ should reflect the long-term motion trend embedded in video, such as motion type (running or walking), direction, etc. Therefore, $\bm{z}_1$ is inferred from the entire pixel frame sequence $\bm{x}_{1:T}$, and the latent variables $\bm{z}_t$ ($2\le t\le T$) are encouraged to reflect inter-frame dynamics in the video sequence. The local motion trend depends on the current frame and previous frames, and similar to prior works ~\cite{franceschi2020stochastic,denton2018stochastic,akan2021slamp}, $\bm{z}_t$ is inferred from the frame sequence $\bm{x}_{1:t}$, while the initial motion variable $\bm{y}_1$ are inferred from the initial two frames. To fit the true prior distribution of the latent variables as closely as possible, we employ a deep variational inference model to simulate the distributions of various latent variables. The overall inference model is designed as shown in Figure~\ref{fig:generativeAndInference}(b), and the variational distribution obtained from the inference model is:
\begin{equation}
    \begin{split}
    q_{Z,Y}=q(\bm{z}_{1:T},&\bm{y}_{1:T}|\bm{x}_{1:T},\bm{w}_{1:T})=q(\bm{z}_1|\bm{x}_{1:T})q(\bm{y}_1|\bm{x}_{1:2})\\
    &\textstyle\prod_{t=2}^{T}q(\bm{z}_t|\bm{x}_{1:t})q(\bm{y}_t|\bm{y}_{_{t-1}},\bm{z}_t,\bm{z}_1) 
    \end{split}
\label{equa: variational distribution}
\end{equation}
Combining the variational distributions from the inference model, the lower bound of the likelihood for the pixel frame sequence $\bm{x}_{1:T}$ can be obtained (complete derivation is provided in Appendix A):
\begin{equation}
    \tablestyle{6pt}{1}\begin{split}
        \log_{}&{p(\bm{x}_{1:T}|\bm{w})} =\int\limits_{\bm{z}} \int\limits_{\bm{y}} q(\bm{z}_{1:T},\bm{y}_{1:T}|\bm{x}_{1:T},\bm{w})\log p(\bm{x}_{1:T}|\bm{w})\mathrm{d}_{\bm{z}}\mathrm{d}_{\bm{y}}\\
        =&\mathbb{E}_{(\bm{z}_{1:T},\bm{y}_{1:T})\sim q_{Z,Y}} [\log p(\bm{x}_t|\bm{y}_t,\bm{w}_t)]-D_{KL}(q(\bm{y}_1|\bm{x}_{1:2})||p(\bm{y}_1)) \\
        &- \mathbb{E}_{(\bm{z}_{1:T},\bm{y}_{1:T})\sim q_{Z,Y}}\sum_{t=2}^{T} D_{KL}[q(\bm{z}_t|\bm{x}_{1:t})||p(\bm{z}_t|\bm{y}_{t-1})]\\
        &-D_{KL}(q(\bm{z}_1|\bm{x}_{1:T})||p(\bm{z}_1|\bm{x}_{1:k}))
    \end{split}
\label{equa: ELBO}
\end{equation}
Here, $D_{KL}$ represents the KL divergence \cite{kullback1951information}. For the initial motion variable, we adopt a strong Gaussian prior aiming to encourage the motion variables to discard unnecessary information. For optimizing the log-likelihood $\log p(\bm{x}_t|\bm{y}_t,\bm{w}_t)$, we compute the gradient by calculating the negative log density function of $\bm{\hat{x}}_t$ with respect to a normal distribution created by ground truth $\bm{x}_t$. For the remaining KL divergence terms, we compute the gradient using the reparameterization technique \cite{kingma2013auto}. For more training details, please refer to Appendix B.

\subsection{Architecture}
The overall framework of our method is illustrated in Figure~\ref{fig:framework}. In the stochastic motion prediction branch, to infer the local dynamics $\bm{z}_t$, we initially employ a convolutional neural network to encode frames into $\bm{h}_{1:T}^m$, i.e., $\bm{h}_{1:T}^{m} =\mathrm{Enc}(\bm{x}_{1:T})$. Subsequently, we use a LSTM to infer the posterior of $z_t$ in a feed-forward fashion:
\begin{equation}
\begin{split}
    \bm{g}_t&=\mathrm{LSTM}_{\beta}(\bm{h}_{1:t}^m)) \\
    \mu_{\beta}(t),&\sigma_{\beta}(t)=\mathrm{MLP_{}} (\bm{g}_{t}^{})
\end{split}
\label{equa: infer posterior of z_t}
\end{equation}
As described in Section 3.1, we derive the prior of local dynamic $\bm{z}_t$ from the previous motion variable, denoted as :
\begin{equation}
\begin{split}
    \mu_{\alpha}(t),\sigma_{\alpha }(t)=\mathrm{MLP}(\bm{y}_{t-1})
\end{split}
\end{equation}
For the initial motion variable $\bm{y}_1$, we infer it using the first two frames, denoted as $\mu_{\psi },\sigma_{\psi }=\mathrm{MLP}(\bm{h}_{1:2}^m)$. Regarding the global dynamic $\bm{z}_1$, our perspective is that overall motion trend, type must be inferred through the complete long-term sequence. Therefore, we infer the posterior of $\bm{z}_1$ using the complete sequence with a temporal transfomer. During testing, when the complete sequence is not visible, we use the conditional frames to generate the prior of $\bm{z}_1$:
\begin{equation}
\begin{split}
    &\mu_{\varphi  },\sigma_{\varphi  }=\mathrm{Transformer_{\varphi}}(\bm{h}_{1:T}^m)\\
    &\mu_{\lambda },\sigma_{\lambda }=\mathrm{Transformer_{\varphi}}(\bm{h}_{1:k}^m)
\end{split}
\end{equation}
For the deterministic appearance prediction branch, previous works \cite{arnab2021vivit,ye2023video} demonstrate the advantages of Vision Transformer (ViT) in adaptively extracting image features. We employ a ViT as the appearance encoder to encode features related to the background. Building upon the strong Gaussian prior for motion variables to discard unnecessary information for motion, the learnable appearance token encourages ViT to adaptively focus on pixel patches related to the background, i.e., $\bm{h}_{t}^{w}=\mathrm{ViT}(\bm{x}_{t})$.
where $\bm{h}_{t}^{w}$ is the static features encoded by ViT, then a LSTM is used to generate inter-frame transition dynamics of
appearance features, i.e., $\tilde{\bm{z}}_t^w =\mathrm{LSTM}_{\vartheta }(\bm{h}_{1:t}^w)$. During testing, $\bm{z}_{t}^{w}$ is predicted from the previous appearance variable. During training, $\tilde{\bm{z}}_{t}^{w}$ is used to supervise $\bm{z}_{t}^{w}$ through $\mathcal{L}_2$ loss $\textstyle \sum_{t=2}^{T} \left \| \tilde{\bm{z}}_t^w-\bm{z}_t^w \right \|_2$. Additionally, to facilitate $\bm{z}_{2:T}$ capturing local inter-frame dynamics, $\mathrm{FlowDec}$ with the same architecture as $Dec$ is employed to decode optical flow $\bm{f}_{2:T}$ from the output $\bm{g}_{2:T}$ of the $\mathrm{LSTM}_{\beta}$ and warp it into frame $\tilde{\bm{x}}_{2:T}$ via differentiable warping \cite{jaderberg2015spatial}. 
\begin{equation}
\tablestyle{5.5cm}{1}\begin{split}
    &\bm{f}(t) = \mathrm{FlowDec} (\bm{g}_t) \\
    &\tilde{\bm{x}}_t=\mathrm{warp}(\bm{f}(t),\bm{x}_{t-1})
\end{split}
\end{equation}
where FlowDec is trained using $\mathcal{L}_2$ loss $ {\textstyle \sum_{t=2}^{T}} \left \| \tilde{\bm{x}}_t -\bm{x}_t\right \|_2$. Please note that predicting optical flow is solely intended to encourage $\bm{z}_t$ to focus on the inter-frame local dynamic of frame $\bm{x}_t$ with respect to $\bm{x}_{t-1}$ during training. The model does not require optical flow to generate future frames.

\begin{figure*}[t]
  \begin{center}
     \includegraphics[width=\linewidth]{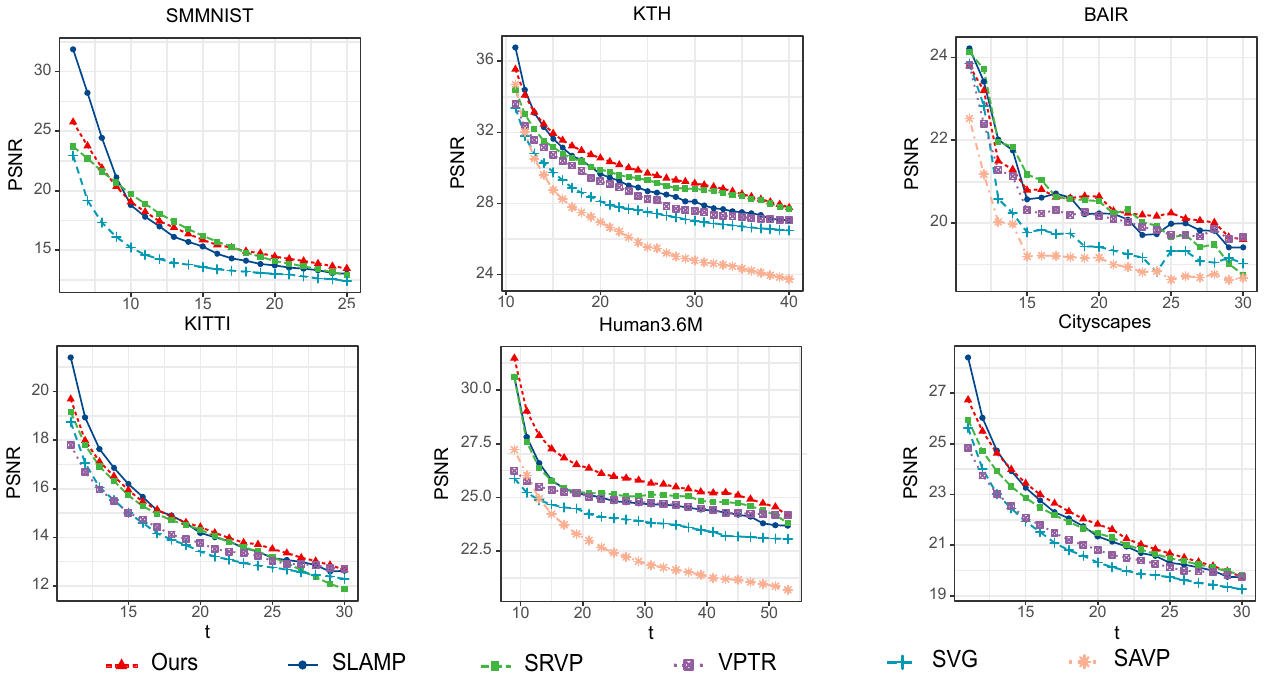}
  \end{center}
     \caption{The PSNR scores over timestep t for our proposed method and various baselines. Each score represents the mean value obtained from five different samples generated by the models. 
     Our proposed model achieved superior performance on the KTH, Human3.6M and Cityscapes datasets, while demonstrating comparable performance to state-of-the-art models on the BAIR, SMMNIST and KITTI datasets in terms of the PSNR metric.}
  \label{fig:line}
\end{figure*}

\section{Experiments}

\subsection{Datasets}
To evaluate the proposed method's effectiveness across diverse scenarios, experiments are conducted on datasets with both stationary and dynamic backgrounds, 
including SMMNIST \cite{denton2018stochastic}, BAIR \cite{ebert2017self}, KTH \cite{schuldt2004recognizing}, Human3.6M \cite{ionescu2011latent}, Cityscapes \cite{cordts2016cityscapes}, and KITTI \cite{geiger2012we}. SMMNIST involves moving two MNIST digits \cite{lecun1998gradient} with possible overlap. The BAIR Push dataset features a robotic arm pushing an object to induce movement. The KTH dataset encompasses human motion videos with various patterns like walking and running. The Human3.6M dataset includes videos of subjects performing actions indoors, exhibiting more variability than KTH. For assessing performance in complex real-world driving scenarios with moving backgrounds, we also evaluate our method on KITTI and Cityscapes datasets. KITTI footage was captured while driving in Karlsruhe, and Cityscapes includes street driving videos from multiple cities, offering more diverse driving environments than KITTI. More details about the datasets are provided in Appendix C.

\begin{table*}[t!]
    \centering
    \resizebox{17cm}{!}{
    \begin{tabular}{l|ccc|ccc}
        \thickhline
        \multirow{2}{*}{Model} &  \multicolumn{3}{c|}{dataset:\textbf{SMMNIST}(5$\to$20)}  & \multicolumn{3}{c}{dataset:\textbf{BAIR}(10$\to$20)}      \\
          & PSNR ($\uparrow$) & SSIM ($\uparrow$)   & LPIPS ($\downarrow$)  & PSNR ($\uparrow$) & SSIM ($\uparrow$)   & LPIPS ($\downarrow$)  \\
        \hline
        SVG   & 14.50$\pm$0.04   & 0.7090$\pm$0.0015 & -  & 19.85$\pm$0.26   & 0.8301$\pm$0.0088  & 0.0553$\pm$0.0034  \\
        SAVP  & -                & -                 & -  & 19.39$\pm$0.25   & 0.8137$\pm$0.0092  & 0.0564$\pm$0.0026  \\
        VPTR &  - & - & -  & 20.41$\pm$0.27   & 0.8363$\pm$0.0088  & 0.0560$\pm$0.0036  \\
        SRVP  & 16.93$\pm$0.07   & 0.7799$\pm$0.0020 & -  &  20.63$\pm$0.28  & \underline{0.8448$\pm$0.0078}  & \underline{0.0521$\pm$0.0032}  \\
        SLAMP & \textbf{18.07$\pm$0.08}   & 0.7736$\pm$0.0019 & -  & \textbf{20.71$\pm$0.26}   & 0.8402$\pm$0.0086  & 0.0575$\pm$0.0037  \\
        Ours  & \underline{17.12$\pm$0.07}     & \textbf{0.7809$\pm$0.0020}       &  -  & \underline{20.67$\pm$0.28}                 &\textbf{0.8459$\pm$0.0078}     &\textbf{0.0520$\pm$0.0032}    \\
        \hline

        \multirow{2}{*}{Model} &  \multicolumn{3}{c|}{dataset:\textbf{KTH}(10$\to$30)}  & \multicolumn{3}{c}{dataset:\textbf{Human3.6M}(8$\to$45)}      \\
          & PSNR ($\uparrow$) & SSIM ($\uparrow$)   & LPIPS ($\downarrow$)  & PSNR ($\uparrow$) & SSIM ($\uparrow$)   & LPIPS ($\downarrow$)  \\
        \hline
        SVG   & 28.06$\pm$0.29 &  0.8438$\pm$0.0054 & 0.0923$\pm$0.0038  & 23.94$\pm$0.19 & 0.8889$\pm$0.0028 &   
        0.0636$\pm$0.0018  \\
        SAVP  & 26.51$\pm$0.29 &  0.7564$\pm$0.0062 & 0.1120$\pm$0.0039  & 22.61$\pm$0.18 & 0.8036$\pm$0.0031 &      0.0764$\pm$0.0019  \\
        VPTR & 28.77$\pm$0.31  & 0.8674$\pm$0.0055 & 0.0851$\pm$0.0035  & 24.82$\pm$0.21   & 0.8948$\pm$0.0026  & 0.0621$\pm$0.0018  \\
        SRVP  & \underline{29.69$\pm$0.32} & \underline{0.8697$\pm$0.0046}  & \textbf{0.0736$\pm$0.0029}  & \underline{25.30$\pm$0.19} & \underline{0.9074$\pm$0.0022} & \underline{0.0509$\pm$0.0013}\\
        SLAMP & 29.39$\pm$0.30 & 0.8646$\pm$0.0050  & 0.0795$\pm$0.0034  & 25.17$\pm$0.19 & 0.9032$\pm$0.0022 &      
        0.0549$\pm$0.0015 \\

        Ours  & \textbf{30.30$\pm$0.31} & \textbf{0.8766$\pm$0.0045}  & \underline{0.0743$\pm$0.0029}  & \textbf{26.07$\pm$0.20} & \textbf{0.9160$\pm$0.0021} & \textbf{0.0501$\pm$0.0013}  \\
        \hline
        
        \multirow{2}{*}{Model} &  \multicolumn{3}{c|}{dataset:\textbf{KITTI}(10$\to$20)}  & \multicolumn{3}{c}{dataset:\textbf{Cityscapes}(10$\to$20)}      \\
          & PSNR ($\uparrow$) & SSIM ($\uparrow$)   & LPIPS ($\downarrow$)  & PSNR ($\uparrow$) & SSIM ($\uparrow$)   & LPIPS ($\downarrow$)  \\
        \hline
        SVG   &13.97$\pm$0.47 &0.3572$\pm$0.0183  &0.5537$\pm$0.0379   &20.94$\pm$0.61  & 0.6211$\pm$0.0218 &0.3094$\pm$0.0209       \\
        VPTR  &14.13$\pm$0.44  &0.3558$\pm$0.0198  &0.5438$\pm$0.0243   &21.24$\pm$0.53  &0.6279$\pm$0.0221  &0.3214$\pm$0.0235       \\
        SRVP  &14.53$\pm$0.34  &0.3637$\pm$0.0195  &0.5264$\pm$0.0235   &21.77$\pm$0.44  &0.6349$\pm$0.0161  &0.3147$\pm$0.0145       \\
        SLAMP &\textbf{14.87$\pm$0.49}  &\underline{0.3698$\pm$0.0207}  &\underline{0.4912$\pm$0.0397}   &\underline{22.01$\pm$0.71}  &\underline{0.6513$\pm$0.0232}  &\textbf{0.2937$\pm$0.0214}       \\
        Ours  &\underline{14.67$\pm$0.46}  &\textbf{0.3781$\pm$0.0230}  &\textbf{0.4572$\pm$0.0236}   &\textbf{22.12$\pm$0.46}  &\textbf{0.6555$\pm$0.0163}  &\underline{0.3014$\pm$0.0134}       \\
        \thickhline
    \end{tabular}
    }
    \caption{Numerical results (mean and 95\%-confidence interval) for PSNR, SSIM, and LPIPS of our proposed method and baselines.}
    \label{tab: com-baselines}
\end{table*}

\subsection{Implementation Details}
For the BAIR, KTH, SMMNIST, and Human3.6M datasets, image frames are 64×64 pixels, and for KITTI and Cityscapes datasets, frames are 128×128 pixels. During training, BAIR, KITTI, Cityscapes, and KTH use the initial 10 frames as conditionals for predicting the next 10 frames; Human3.6M uses the initial 8 frames to predict the subsequent 8 frames; SMMNIST uses the initial 5 frames to predict the next 10 frames. During testing, BAIR, Cityscapes, KITTI, and SMMNIST predict the next 20 frames. For KTH, 30 frames are predicted, and for Human3.6M, 45 frames are predicted. Our encoder-decoder adopts the VGG16 \cite{simonyan2014very} architecture for a fair comparison with the previous methods ~\cite{franceschi2020stochastic,akan2021slamp}. Additional training details are available in Appendix B. \\
\textbf{Baselines and Evaluation Metrics}:
To evaluate our state-space decomposition model for stochastic video prediction, we compare it with leading variational methods (SVG \cite{denton2018stochastic}, SAVP \cite{lee2018stochastic}, SRVP \cite{franceschi2020stochastic}, SLAMP \cite{akan2021slamp}) and a deterministic approach (VPTR \cite{ye2023video}) on various datasets. We use three standard metrics for future frame prediction performance: Peak Signal-to-Noise Ratio (PSNR) for reconstruction quality, Structural Similarity Index (SSIM) for structural alignment, and Learned Perceptual Image Patch Similarity (LPIPS \cite{zhang2018unreasonable}) for dissimilarity assessment based on feature maps.
\begin{figure}[!t]
  \begin{center}
     \includegraphics[width=\linewidth]{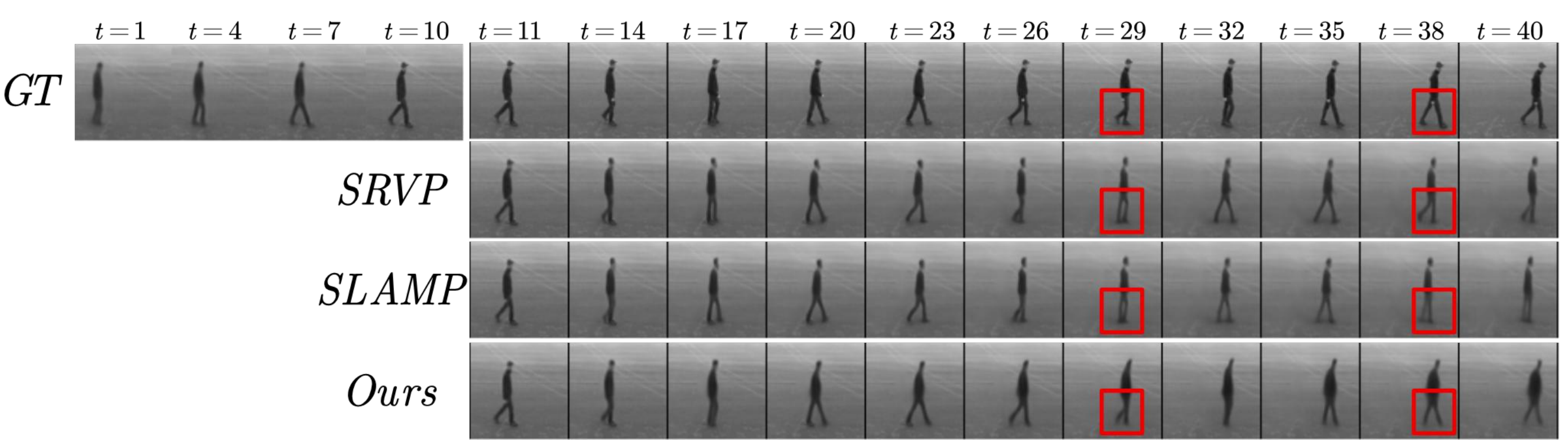}
  \end{center}
     \caption{Person walking. The top row shows the ground truth, followed by the predictions from SRVP, SLAMP and our method.}
  \label{fig:KTH1}
\end{figure}
\begin{figure}[!t]
  \begin{center}
     \includegraphics[width=\linewidth]{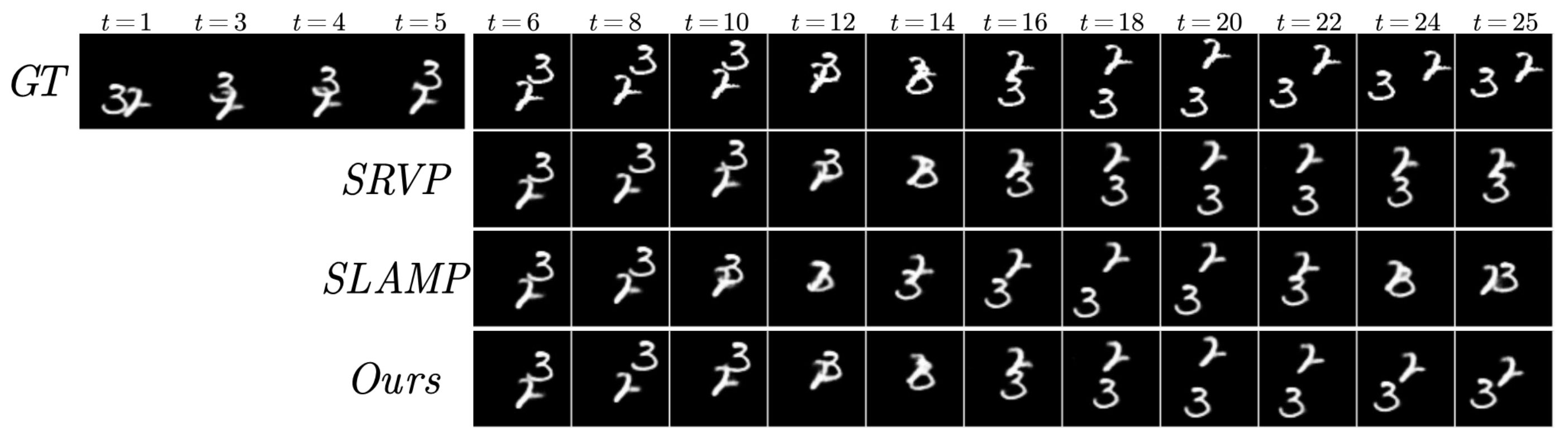}
  \end{center}
     \caption{Overlapping digits. This figure shows two overlapping digits and the predictions from SRVP, SLAMP and our method.}
  \label{fig:SMMNIST3}
\end{figure}
\subsection{Evaluation}
\begin{figure}[!t]
  \begin{center}
     \includegraphics[width=\linewidth]{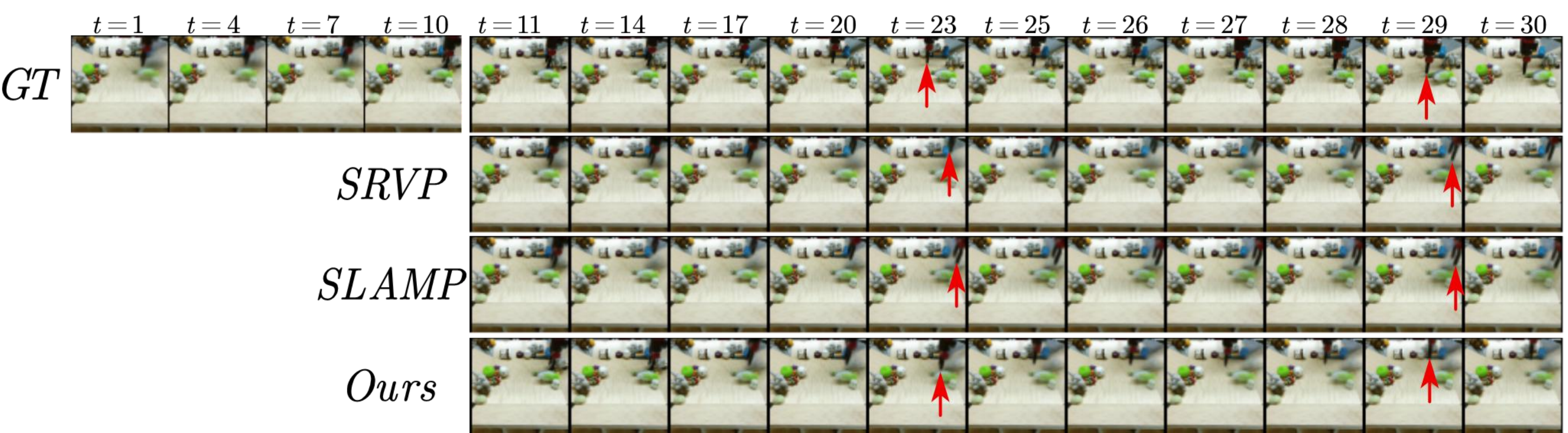}
  \end{center}
     \caption{Results on BAIR. Our method accurately predicts the long-term displacement of the robotic arm compared to baselines.}
  \label{fig:BAIR1}
\end{figure}

Table \ref{tab: com-baselines} presents quantitative results of our method and baselines across various datasets. Our method achieves state-of-the-art (SOTA) performance across multiple datasets, particularly excelling in the SSIM metric. In KTH dataset, our method demonstrates superior performance in inferring the long-term motion context coherently with the conditional frames, as illustrated in Figure \ref{fig:KTH1}, in the case of human walking samples, our method accepts constraints from global dynamics, leading to more realistic leg details in the predictions. Specifically, on the Human3.6M dataset, where only 8 frames are used to predict the next 45 frames, our method outperforms baselines on all three metrics. The KITTI and Cityscapes datasets contain real driving videos with changing backgrounds over time. As indicated in Table \ref{tab: com-baselines}, our method outperforms SRVP and SVG significantly on KITTI and Cityscapes datasets, achieving a level comparable to SLAMP while maintaining a computational advantage (see Table \ref{tab:ablation}). For SMMNIST dataset, the challenge lies in the separation prediction of intersecting digits. As shown in Figure \ref{fig:SMMNIST3}, our method benefits from the guidance of the global motion trend $\bm{z}_1$, successfully decoupling two digits even after their intersection, and accurately predicting the motion direction of each digit. On the BAIR dataset, our method surpasses baseline methods in both PSNR and SSIM metrics, though it falls behind SRVP in terms of LPIPS. When predicting the long-term future, as shown in Figure \ref{fig:BAIR1}, our method accurately predicts the long-term displacement of the robotic arm.

\begin{figure}[!t]
  \begin{center}
     \includegraphics[width=\linewidth]{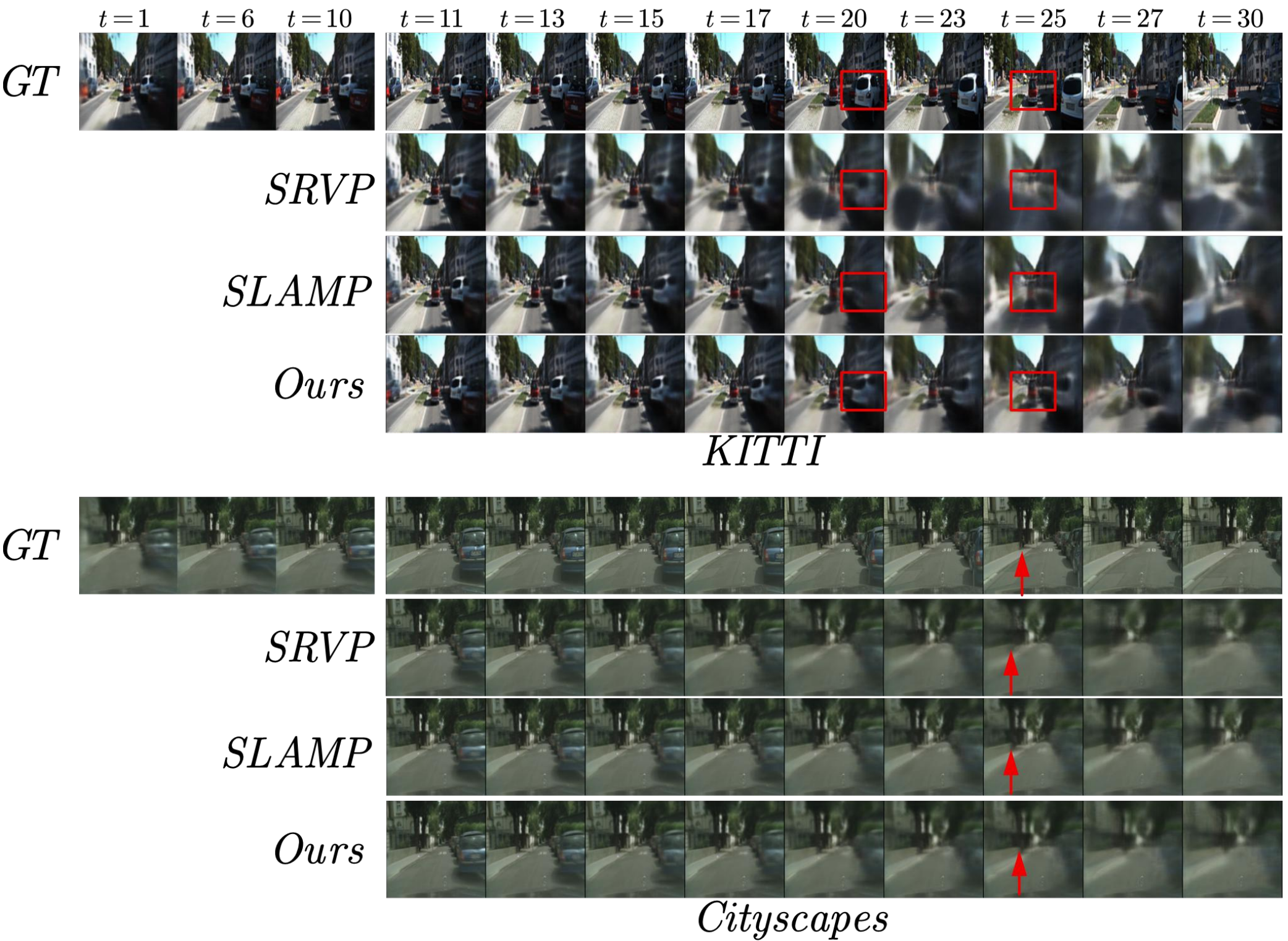}
  \end{center}
     \caption{Results on KITTI (\textbf{up}) and Cityscapes (\textbf{down}). Our method predicts richer details (red square) and achieves more accurate positioning of the street trees (red arrow).}
  \label{fig:kittiAndCity}
\end{figure}

In order to contrast the performance of our method and baselines at each time step, we plot the relationship between prediction quality and time steps in Figure \ref{fig:line}. It cant be seen that on the KTH and Human3.6M datasets, the proposed method consistently outperforms baseline methods at each step. On the challenging KITTI and Cityscapes datasets, our method initially lags behind the state-of-the-art method SLAMP in the first few steps. However, it gradually surpasses SLAMP in later steps, owing to the decoupling of frame synthesis and temporal prediction by the state space model, resulting in smoother sequence predictions. Specific sequence visualizations are illustrated in Figure \ref{fig:kittiAndCity}. For SMMIST, SLAMP exhibits superior performance in the initial steps but experiences a rapid decline, falling short of SRVP and our method. In the BAIR dataset, our method achieves superior performance relative to baselines after step 20. Featuring a rapidly moving robotic arm with continuously changing directions, all the models have noticeable discontinuities.

\subsection{Experiments on State-space Decomposition}
\begin{figure}[!t]
  \begin{center}
     \includegraphics[width=\linewidth]{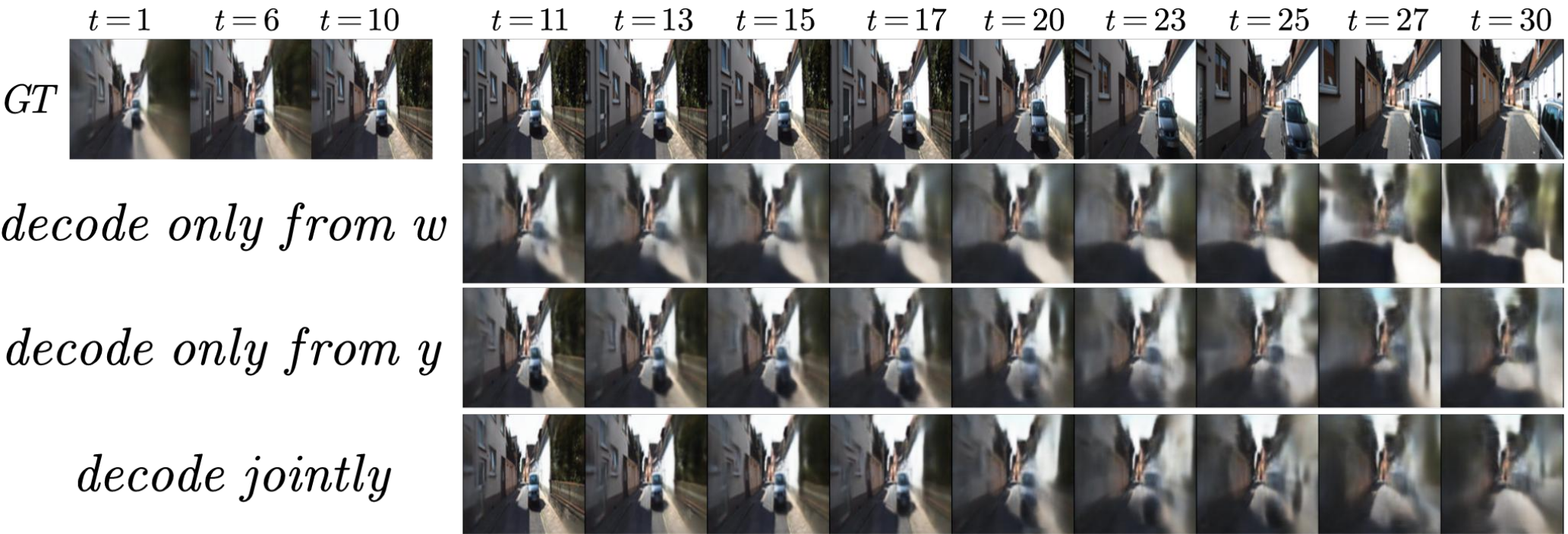}
  \end{center}
     \caption{Results decoded separately. This figure shows videos decoded separately from $w$ and $y$, and the result of joint decoding.}
  \label{fig:onlyW and onlyY}
\end{figure}

\begin{figure}[!t]
  \begin{center}
     \includegraphics[width=\linewidth]{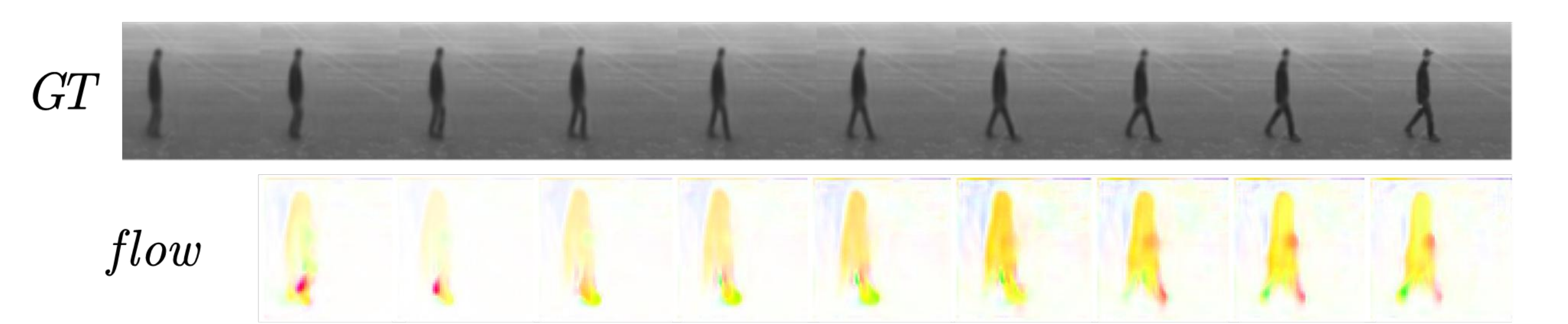}
  \end{center}
     \caption{Oplical flow. This figure shows the pixel sequence and the opliacal flow decoded from the output of $\mathrm{LSTM}_{\beta}$.}
  \label{fig:flow}
\end{figure}

Our method decomposes video prediction into stochastic motion prediction and deterministic appearance prediction. The Gaussian prior on the initial motion variable facilitates the attention of motion variables on the dynamic features of the subject. The learnable appearance token encourages the ViT to focus on static features related to the background. 
To validate the performance of different branches, we visualize sequences decoded separately from the appearance variable $\bm{w}$ and the motion variable $\bm{y}$ in Figure \ref{fig:onlyW and onlyY}. It can be observed that the motion variables capture the movement of the subject (i.e., the car), including the motion direction and displacement, while appearance variables focuses more on static features such as background contours and spatial relationships. To further verify the ability of the stochastic motion prediction branch to model inter-frame local dynamics, we visualize the optical flow decoded from the output of $\mathrm{LSTM}_{\beta}$ in Figure \ref{fig:flow}. $\mathrm{LSTM}_{\beta}$ effectively captures local dynamics, demonstrating the rationality of predicting the distribution of local motion trend $z_{2:T}$ based on the output of $\mathrm{LSTM}_{\beta}$ in equation \ref{equa: infer posterior of z_t}. Utilizing the ViT with a learnable appearance token allows capturing static information in the frame sequence.
We also compare the case where VGG16 serves as the appearance encoder in Appendix D.


\subsection{Ablation Studies}
To further validate the impact of the appearance prediction branch and the global dynamics, we compare the performance in the following settings: (\textbf{i}) Ours (w/o $\bm{w}$), abandoning the appearance prediction branch and adopting the same static content variable scheme as SRVP \cite{franceschi2020stochastic}, (\textbf{ii}) Ours (w/o $\bm{z}_1$), Not considering the global dynamic $\bm{z}_1$ when predicting frame-to-frame transitions in the motion prediction branch, and (\textbf{iii}) Ours method, as described in Chapter 3. The results are presented in Table \ref{tab:ablation}, where each component contributes to the predictive performance. These results indicate the effectiveness of the deterministic appearance prediction branch in adaptively encoding images and the global dynamic $\bm{z}_1$ for predicting long-term frame transitions. Additionally, the computational time increase introduced by the added components is marginal compared to SRVP. For more ablation experiments and visualization samples, please refer to Appendix D.

\begin{table}[!htbp]
    \centering
    {
    \tablestyle{6pt}{1}\begin{tabular}{c||c|c|c|c}
    \thickhline
    \rowcolor{mygray}
    \rowcolor{mygray}
    \multirow{-1}*{Method} & \multicolumn{1}{c}{PSNR$\uparrow$} & \multicolumn{1}{c}{SSIM$\uparrow$} & \multicolumn{1}{c}{LPIPS$\downarrow$} & \multicolumn{1}{c}{Inf Time(s)$\downarrow$}  \\ \hline\hline
    SRVP & 14.53 & 0.3637 & 0.5264 & \textbf{0.035}  \\
    SLAMP & \textbf{14.87} & 0.3698 & 0.4912 & 0.369 \\
    Ours (w/o $w$) & \underline{14.79} & 0.3655 &0.5068  & \underline{0.043} \\
    Ours (w/o $z_1$) & 14.50 & \underline{0.3723} & \underline{0.4731} & 0.049 \\
    Ours  & 14.67 & \textbf{0.3781} & \textbf{0.4572} & 0.058 \\
    \thickhline
    \end{tabular}
    }
    \caption{Ablation results on KITTI regarding PSNR, SSIM, LPIPS, and Inference Time (average inference time for testing 100 samples on RTX 2080Ti).}
    \label{tab:ablation}
\end{table}

\section{Conclusion}
In this paper, we propose a state space decomposition video prediction model that decomposes the overall frame prediction into stochastic motion prediction and deterministic appearance prediction.
The stochastic motion prediction module, when predicting inter-frame residuals, incorporates the global dynamics extracted from the conditional sequence to guide the motion predictions. Experimental results demonstrate that the proposed method achieves state-of-the-art performance on various datasets. 

\newpage

\bibliographystyle{named}
\bibliography{ijcai24}

\begin{thebibliography}{}

\bibitem[\protect\citeauthoryear{Akan \bgroup \em et al.\egroup }{2021}]{akan2021slamp}
Adil~Kaan Akan, Erkut Erdem, Aykut Erdem, and Fatma G{\"u}ney.
\newblock Slamp: Stochastic latent appearance and motion prediction.
\newblock In {\em Proceedings of the IEEE/CVF International Conference on Computer Vision}, pages 14728--14737, 2021.

\bibitem[\protect\citeauthoryear{Arnab \bgroup \em et al.\egroup }{2021}]{arnab2021vivit}
Anurag Arnab, Mostafa Dehghani, Georg Heigold, Chen Sun, Mario Lu{\v{c}}i{\'c}, and Cordelia Schmid.
\newblock Vivit: A video vision transformer.
\newblock In {\em Proceedings of the IEEE/CVF International Conference on Computer Vision}, pages 6836--6846, 2021.

\bibitem[\protect\citeauthoryear{Chang \bgroup \em et al.\egroup }{2022}]{chang2022strpm}
Zheng Chang, Xinfeng Zhang, Shanshe Wang, Siwei Ma, and Wen Gao.
\newblock Strpm: A spatiotemporal residual predictive model for high-resolution video prediction.
\newblock In {\em Proceedings of the IEEE/CVF Conference on Computer Vision and Pattern Recognition}, pages 13946--13955, 2022.

\bibitem[\protect\citeauthoryear{Chatterjee \bgroup \em et al.\egroup }{2021}]{chatterjee2021hierarchical}
Moitreya Chatterjee, Narendra Ahuja, and Anoop Cherian.
\newblock A hierarchical variational neural uncertainty model for stochastic video prediction.
\newblock In {\em Proceedings of the IEEE/CVF International Conference on Computer Vision}, pages 9751--9761, 2021.

\bibitem[\protect\citeauthoryear{Cordts \bgroup \em et al.\egroup }{2016}]{cordts2016cityscapes}
Marius Cordts, Mohamed Omran, Sebastian Ramos, Timo Rehfeld, Markus Enzweiler, Rodrigo Benenson, Uwe Franke, Stefan Roth, and Bernt Schiele.
\newblock The cityscapes dataset for semantic urban scene understanding.
\newblock In {\em Proceedings of the IEEE Conference on Computer Vision and Pattern Recognition}, pages 3213--3223, 2016.

\bibitem[\protect\citeauthoryear{Denton and Birodkar}{2017}]{denton2017unsupervised}
Emily Denton and Vighnesh Birodkar.
\newblock Unsupervised learning of disentangled representations from video.
\newblock {\em Advances in Neural Information Processing Systems}, 30:1--10, 2017.

\bibitem[\protect\citeauthoryear{Denton and Fergus}{2018}]{denton2018stochastic}
Emily Denton and Rob Fergus.
\newblock Stochastic video generation with a learned prior.
\newblock In {\em International Conference on Machine Learning}, pages 1174--1183, 2018.

\bibitem[\protect\citeauthoryear{Dugas \bgroup \em et al.\egroup }{2022}]{dugas2022navdreams}
Daniel Dugas, Olov Andersson, Roland Siegwart, and Jen~Jen Chung.
\newblock Navdreams: Towards camera-only rl navigation among humans.
\newblock In {\em 2022 IEEE/RSJ International Conference on Intelligent Robots and Systems}, pages 2504--2511, 2022.

\bibitem[\protect\citeauthoryear{Ebert \bgroup \em et al.\egroup }{2017}]{ebert2017self}
Frederik Ebert, Chelsea Finn, Alex~X Lee, and Sergey Levine.
\newblock Self-supervised visual planning with temporal skip connections.
\newblock {\em CoRL}, 12:16, 2017.

\bibitem[\protect\citeauthoryear{Farazi \bgroup \em et al.\egroup }{2021}]{farazi2021local}
Hafez Farazi, Jan Nogga, and Sven Behnke.
\newblock Local frequency domain transformer networks for video prediction.
\newblock In {\em 2021 International Joint Conference on Neural Networks}, pages 1--10, 2021.

\bibitem[\protect\citeauthoryear{Finn and Levine}{2017}]{finn2017deep}
Chelsea Finn and Sergey Levine.
\newblock Deep visual foresight for planning robot motion.
\newblock In {\em 2017 IEEE International Conference on Robotics and Automation}, pages 2786--2793, 2017.

\bibitem[\protect\citeauthoryear{Franceschi \bgroup \em et al.\egroup }{2020}]{franceschi2020stochastic}
Jean-Yves Franceschi, Edouard Delasalles, Micka{\"e}l Chen, Sylvain Lamprier, and Patrick Gallinari.
\newblock Stochastic latent residual video prediction.
\newblock In {\em International Conference on Machine Learning}, pages 3233--3246, 2020.

\bibitem[\protect\citeauthoryear{Gao \bgroup \em et al.\egroup }{2019}]{gao2019disentangling}
Hang Gao, Huazhe Xu, Qi-Zhi Cai, Ruth Wang, Fisher Yu, and Trevor Darrell.
\newblock Disentangling propagation and generation for video prediction.
\newblock In {\em Proceedings of the IEEE/CVF International Conference on Computer Vision}, pages 9006--9015, 2019.

\bibitem[\protect\citeauthoryear{Geiger \bgroup \em et al.\egroup }{2012}]{geiger2012we}
Andreas Geiger, Philip Lenz, and Raquel Urtasun.
\newblock Are we ready for autonomous driving? the kitti vision benchmark suite.
\newblock In {\em 2012 IEEE Conference on Computer Vision and Pattern Recognition}, pages 3354--3361, 2012.

\bibitem[\protect\citeauthoryear{Goel \bgroup \em et al.\egroup }{2022}]{goel2022s}
Karan Goel, Albert Gu, Chris Donahue, and Christopher R{\'e}.
\newblock It’s raw! audio generation with state-space models.
\newblock In {\em International Conference on Machine Learning}, pages 7616--7633, 2022.

\bibitem[\protect\citeauthoryear{Gregor \bgroup \em et al.\egroup }{2018}]{gregor2018temporal}
Karol Gregor, George Papamakarios, Frederic Besse, Lars Buesing, and Theophane Weber.
\newblock Temporal difference variational auto-encoder.
\newblock {\em arXiv Preprint arXiv:1806.03107}, 2018.

\bibitem[\protect\citeauthoryear{Hafner \bgroup \em et al.\egroup }{2019}]{hafner2019learning}
Danijar Hafner, Timothy Lillicrap, Ian Fischer, Ruben Villegas, David Ha, Honglak Lee, and James Davidson.
\newblock Learning latent dynamics for planning from pixels.
\newblock In {\em International Conference on Machine Learning}, pages 2555--2565, 2019.

\bibitem[\protect\citeauthoryear{Ionescu \bgroup \em et al.\egroup }{2011}]{ionescu2011latent}
Catalin Ionescu, Fuxin Li, and Cristian Sminchisescu.
\newblock Latent structured models for human pose estimation.
\newblock In {\em 2011 International Conference on Computer Vision}, pages 2220--2227, 2011.

\bibitem[\protect\citeauthoryear{Jaderberg \bgroup \em et al.\egroup }{2015}]{jaderberg2015spatial}
Max Jaderberg, Karen Simonyan, Andrew Zisserman, et~al.
\newblock Spatial transformer networks.
\newblock {\em Advances in Neural Information Processing Systems}, 28, 2015.

\bibitem[\protect\citeauthoryear{Jin \bgroup \em et al.\egroup }{2020}]{jin2020exploring}
Beibei Jin, Yu~Hu, Qiankun Tang, Jingyu Niu, Zhiping Shi, Yinhe Han, and Xiaowei Li.
\newblock Exploring spatial-temporal multi-frequency analysis for high-fidelity and temporal-consistency video prediction.
\newblock In {\em Proceedings of the IEEE/CVF Conference on Computer Vision and Pattern Recognition}, pages 4554--4563, 2020.

\bibitem[\protect\citeauthoryear{Kalchbrenner \bgroup \em et al.\egroup }{2017}]{kalchbrenner2017video}
Nal Kalchbrenner, A{\"a}ron Oord, Karen Simonyan, Ivo Danihelka, Oriol Vinyals, Alex Graves, and Koray Kavukcuoglu.
\newblock Video pixel networks.
\newblock In {\em International Conference on Machine Learning}, pages 1771--1779, 2017.

\bibitem[\protect\citeauthoryear{Karl \bgroup \em et al.\egroup }{2016}]{karl2016deep}
Maximilian Karl, Maximilian Soelch, Justin Bayer, and Patrick Van~der Smagt.
\newblock Deep variational bayes filters: Unsupervised learning of state space models from raw data.
\newblock {\em arXiv Preprint arXiv:1605.06432}, 2016.

\bibitem[\protect\citeauthoryear{Kingma and Welling}{2013}]{kingma2013auto}
Diederik~P Kingma and Max Welling.
\newblock Auto-encoding variational bayes.
\newblock {\em arXiv Preprint arXiv:1312.6114}, 2013.

\bibitem[\protect\citeauthoryear{Kullback and Leibler}{1951}]{kullback1951information}
Solomon Kullback and Richard~A Leibler.
\newblock On information and sufficiency.
\newblock {\em The Annals of Mathematical Statistics}, 22(1):79--86, 1951.

\bibitem[\protect\citeauthoryear{LeCun \bgroup \em et al.\egroup }{1998}]{lecun1998gradient}
Yann LeCun, L{\'e}on Bottou, Yoshua Bengio, and Patrick Haffner.
\newblock Gradient-based learning applied to document recognition.
\newblock {\em Proceedings of the IEEE}, 86(11):2278--2324, 1998.

\bibitem[\protect\citeauthoryear{Lee \bgroup \em et al.\egroup }{2018}]{lee2018stochastic}
Alex~X Lee, Richard Zhang, Frederik Ebert, Pieter Abbeel, Chelsea Finn, and Sergey Levine.
\newblock Stochastic adversarial video prediction.
\newblock {\em arXiv Preprint arXiv:1804.01523}, 2018.

\bibitem[\protect\citeauthoryear{Liang \bgroup \em et al.\egroup }{2017}]{liang2017dual}
Xiaodan Liang, Lisa Lee, Wei Dai, and Eric~P Xing.
\newblock Dual motion gan for future-flow embedded video prediction.
\newblock In {\em proceedings of the IEEE International Conference on Computer Vision}, pages 1744--1752, 2017.

\bibitem[\protect\citeauthoryear{Micheli \bgroup \em et al.\egroup }{2022}]{micheli2022transformers}
Vincent Micheli, Eloi Alonso, and Fran{\c{c}}ois Fleuret.
\newblock Transformers are sample-efficient world models.
\newblock In {\em The Eleventh International Conference on Learning Representations}, pages 1--21, 2022.

\bibitem[\protect\citeauthoryear{Newman \bgroup \em et al.\egroup }{2023}]{newman2023state}
Ken Newman, Ruth King, V{\'\i}ctor Elvira, Perry de~Valpine, Rachel~S McCrea, and Byron~JT Morgan.
\newblock State-space models for ecological time-series data: Practical model-fitting.
\newblock {\em Methods in Ecology and Evolution}, 14(1):26--42, 2023.

\bibitem[\protect\citeauthoryear{Piergiovanni \bgroup \em et al.\egroup }{2019}]{piergiovanni2019learning}
AJ~Piergiovanni, Alan Wu, and Michael~S Ryoo.
\newblock Learning real-world robot policies by dreaming.
\newblock In {\em 2019 IEEE/RSJ International Conference on Intelligent Robots and Systems}, pages 7680--7687, 2019.

\bibitem[\protect\citeauthoryear{Schuldt \bgroup \em et al.\egroup }{2004}]{schuldt2004recognizing}
Christian Schuldt, Ivan Laptev, and Barbara Caputo.
\newblock Recognizing human actions: a local svm approach.
\newblock In {\em Proceedings of the 17th International Conference on Pattern Recognition}, volume~3, pages 32--36, 2004.

\bibitem[\protect\citeauthoryear{Simonyan and Zisserman}{2014}]{simonyan2014very}
Karen Simonyan and Andrew Zisserman.
\newblock Very deep convolutional networks for large-scale image recognition.
\newblock {\em arXiv Preprint arXiv:1409.1556}, 2014.

\bibitem[\protect\citeauthoryear{Vondrick and Torralba}{2017}]{vondrick2017generating}
Carl Vondrick and Antonio Torralba.
\newblock Generating the future with adversarial transformers.
\newblock In {\em Proceedings of the IEEE Conference on Computer Vision and Pattern Recognition}, pages 1020--1028, 2017.

\bibitem[\protect\citeauthoryear{Walker \bgroup \em et al.\egroup }{2016}]{walker2016uncertain}
Jacob Walker, Carl Doersch, Abhinav Gupta, and Martial Hebert.
\newblock An uncertain future: Forecasting from static images using variational autoencoders.
\newblock In {\em Computer Vision--ECCV 2016: 14th European Conference, Amsterdam, The Netherlands, October 11--14, 2016, Proceedings, Part VII 14}, pages 835--851, 2016.

\bibitem[\protect\citeauthoryear{Wang \bgroup \em et al.\egroup }{2017}]{wang2017predrnn}
Yunbo Wang, Mingsheng Long, Jianmin Wang, Zhifeng Gao, and Philip~S Yu.
\newblock Predrnn: Recurrent neural networks for predictive learning using spatiotemporal lstms.
\newblock {\em Advances in Neural Information Processing Systems}, 30:1--10, 2017.

\bibitem[\protect\citeauthoryear{Wang \bgroup \em et al.\egroup }{2018}]{wang2018eidetic}
Yunbo Wang, Lu~Jiang, Ming-Hsuan Yang, Li-Jia Li, Mingsheng Long, and Li~Fei-Fei.
\newblock Eidetic 3d lstm: A model for video prediction and beyond.
\newblock In {\em International Conference on Learning Representations}, pages 1--14, 2018.

\bibitem[\protect\citeauthoryear{Weissenborn \bgroup \em et al.\egroup }{2019}]{weissenborn2019scaling}
Dirk Weissenborn, Oscar T{\"a}ckstr{\"o}m, and Jakob Uszkoreit.
\newblock Scaling autoregressive video models.
\newblock In {\em International Conference on Learning Representations}, pages 1--24, 2019.

\bibitem[\protect\citeauthoryear{Wu \bgroup \em et al.\egroup }{2020}]{wu2020future}
Yue Wu, Rongrong Gao, Jaesik Park, and Qifeng Chen.
\newblock Future video synthesis with object motion prediction.
\newblock In {\em Proceedings of the IEEE/CVF Conference on Computer Vision and Pattern Recognition}, pages 5539--5548, 2020.

\bibitem[\protect\citeauthoryear{Xu \bgroup \em et al.\egroup }{2018}]{xu2018structure}
Jingwei Xu, Bingbing Ni, Zefan Li, Shuo Cheng, and Xiaokang Yang.
\newblock Structure preserving video prediction.
\newblock In {\em Proceedings of the IEEE Conference on Computer Vision and Pattern Recognition}, pages 1460--1469, 2018.

\bibitem[\protect\citeauthoryear{Ye and Bilodeau}{2023}]{ye2023video}
Xi~Ye and Guillaume-Alexandre Bilodeau.
\newblock Video prediction by efficient transformers.
\newblock {\em Image and Vision Computing}, 130:104612, 2023.

\bibitem[\protect\citeauthoryear{Zhang \bgroup \em et al.\egroup }{2018}]{zhang2018unreasonable}
Richard Zhang, Phillip Isola, Alexei~A Efros, Eli Shechtman, and Oliver Wang.
\newblock The unreasonable effectiveness of deep features as a perceptual metric.
\newblock In {\em Proceedings of the IEEE Conference on Computer Vision and Pattern Recognition}, pages 586--595, 2018.

\end{thebibliography}

\end{document}